\documentclass[conference]{IEEEtran}
\IEEEoverridecommandlockouts 
\usepackage[utf8]{inputenc}
\usepackage[T1]{fontenc}
\usepackage{graphicx}
\usepackage{booktabs}
\usepackage[table]{xcolor}
\usepackage{amsmath}
\usepackage{amssymb}
\usepackage[hyphens]{url}
\usepackage{textcomp}

\newif\ifanonymous
\anonymousfalse

\newcommand{\authorA}{John Church}
\newcommand{\orcidA}{https://orcid.org/0000-0002-9472-8101}
\newcommand{\emailA}{johnrobertchurch@protonmail.com}
\newcommand{\authorB}{Vazghen Nikolian}
\newcommand{\orcidB}{https://orcid.org/0009-0003-6337-3496}
\newcommand{\emailB}{nikolainvazghen@gmail.com}

\newcommand{\et}{$\bar{E}_t$}
\newcommand{\eq}{$E_q$}
\newcommand{\epose}{$E_{\mathrm{pose}}$}
\newcommand{\degree}{$^\circ$}

\begin{document}

\title{Vision Foundation Models with Synthetic-Only Training for Monocular Spacecraft Pose Estimation}

\ifanonymous
\author{\IEEEauthorblockN{Anonymous submission}}
\else
\author{%
\IEEEauthorblockN{\authorA\IEEEauthorrefmark{1} \quad \authorB\IEEEauthorrefmark{1}}
\thanks{\IEEEauthorrefmark{1}Both authors contributed equally to this work; author order is alphabetical.}}
\fi

\maketitle
\bstctlcite{IEEEexample:BSTcontrol}

\begin{abstract}
We present an improvement on previous spacecraft pose estimation architectures that results in the lowest published mean rotation errors we know of on the SPEED+ lightbox and sunlamp test sets for a known, non-cooperative spacecraft. By using a previously established heatmap-based pose estimation architecture and adapting a large self-supervised ViT foundation model (DINOv3) in place of the smaller convolutional and ViT encoders of previous work, we show that pose estimation accuracy improves from 300M to 840M parameters with no saturation yet observed. We also evaluate our 840M model on a Jetson Orin NX 16GB, measuring single-pass network inference at 133.8\,ms per crop with a board draw of 32.0\,W. These measurements demonstrate embedded inference feasibility on a processor family with orbital flight heritage. Our resulting model outperforms previous models across lightbox and sunlamp domains while training only on synthetic data. Our best model, using DINOv3 840M adapted with LoRA as the encoder (rank 64, three-seed ensemble with four-rotation test-time augmentation), results in 1.56\degree{} mean rotation error on sunlamp and 1.17\degree{} on lightbox, compared to the previous best mean rotation errors we know of on these test sets, 2.66\degree{} and 1.75\degree{} by EagerNet.
\end{abstract}

\section{Introduction}

In the coming years the number of missions that require a rendezvous between two craft in orbit will likely increase considerably~\cite{isam2025,swf2025rpo,novaspace2026ios}. In the best cases, the target craft in question will be equipped with modern docking features that reduce the risk and complexity of the maneuver. Features like IR dots around docking ports and shared telemetry over stable connections between craft make pose estimation, the problem of an approaching craft knowing the target's 6D pose (relative position and orientation), straightforward. However, many missions will need to be conducted that are not so easy. Targets that are legacy, natural (such as asteroids), broken, or otherwise uncooperative will nonetheless need to be approached and therefore their 6D pose will need to be accurately ascertained without pre-planned cooperative systems assisting.\footnote{Here, non-cooperative refers to the absence of assistance from the target spacecraft; our method still assumes that its geometry and the 3D locations of the selected keypoints are known.}

An additional layer of difficulty is added when considering the constraints imposed by the particular environment the missions will take place in and the material limits of the craft that will be engaging in them. Imagery taken of the target craft is subject to harsh sun lighting that, without the aid of atmospheric diffusion, casts pitch black shadows and overexposes metallic surfaces in direct light. Earth albedo provides some soft fill illumination but cannot be relied upon. Active sensors that would alleviate these visual difficulties somewhat, such as LiDAR, add mass, power draw, and system complexity (the RVS~3000-3D rendezvous LiDAR used for the MEV-1 docking, for example, is specified at 12.4--15.3\,kg and roughly 71\,W nominal power~\cite{rvs3000}). While monocular pose estimation also incurs computational and energy costs, most methods can run on an embedded GPU without requiring a custom processor designed specifically for pose estimation.

\begin{figure}[t]
\centering
\includegraphics[width=\columnwidth]{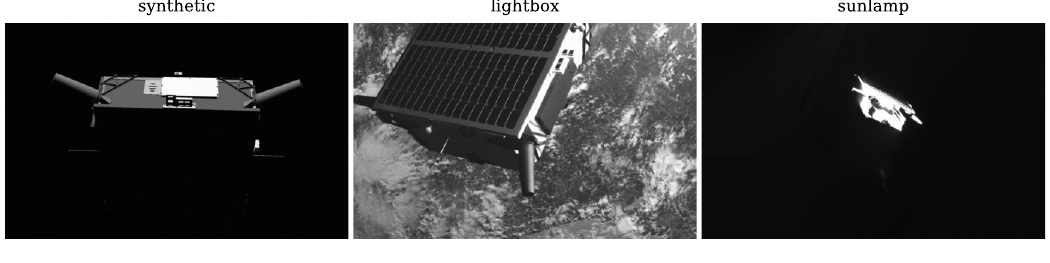}
\caption{Example SPEED+ imagery: synthetic training render (left), lightbox HIL test image (center), sunlamp HIL test image (right). Imagery from SPEED+~\cite{speedplus}, CC BY-NC-SA 4.0.}
\label{fig:examples}
\end{figure}

For these reasons, pose estimation using greyscale monocular imagery has been an object of research interest~\cite{speedplus}. Given the unique difficulty of obtaining real training imagery, the problem of solving pose estimation has gone hand in hand with the problem of bridging the gap between simulated and real training data. SPEED+ was developed as a benchmark by Park et al.~\cite{speedplus}, a collaboration between Stanford's Space Rendezvous Laboratory (SLAB) and ESA's Advanced Concepts Team, and followed by a competition in 2021--2022, the Satellite Pose Estimation Competition (SPEC2021), hosted on ESA's Kelvins platform~\cite{spec2021}. The SPEED+ dataset includes both simulated images of a satellite `Tango' along with real images taken of a physical Tango model staged to simulate harsh direct sun (`sunlamp') as well as earth albedo (`lightbox'). The real imagery comprises 9{,}531 hardware-in-the-loop (HIL) test images of the mockup in Stanford's TRON facility alongside roughly 60{,}000 synthetic images (training and validation), examples of which are shown in Fig.~\ref{fig:examples}.

Significant progress has been made in recent years in solving both the sim to real gap in this domain as well as the central problem of pose estimation itself. The competition was won with mean rotation errors of 2.73\degree{} on sunlamp (team lava1302) and 3.19\degree{} on lightbox (team TangoUnchained)~\cite{spec2021}, though the top entrants all used some of the real imagery during training. The competition postmortem further improved on the competition results with entrants such as FA-VAE~\cite{favae} and the semi-supervised domain adaptation of Chen et al.~\cite{chen2024}. In more recent years, further refinements have pushed the edge further while showing that state of the art results are possible without ever using non-simulated data in training. Impressive results include those of Ulmer et al.'s EagerNet~\cite{eagernet} whose technique also markedly improves on many previous methods by producing state of the art results while requiring only an approximate 3D model of the target. Park and D'Amico, the team behind SPNv3~\cite{spnv3}, showed that it was possible to produce competitive results on a model of only 22.7M parameters (compared to EagerNet's approximately 88.6M-parameter backbone). Concurrently with our research, Bergia et al.~\cite{bergia2026} also implemented a DINOv3 backbone with a heatmap head, obtaining their best result by self-training on unlabeled HIL imagery with DINOv3-L at $224\times224$ input. Our contribution uses low-rank adapters at $448\times448$ input, scales to the 840M encoder, and includes controls that isolate the source of the gains to encoder scale and adaptation. On the same DINOv3-L encoder, our single-pass synthetic-only model reaches 1.90\degree{} / 2.26\degree{} (lightbox / sunlamp) against their 3.27\degree{} / 4.06\degree{}, a 42\% / 44\% reduction in mean rotation error, and our best configuration reaches 1.17\degree{} / 1.56\degree{} against their self-trained 2.57\degree{} / 3.25\degree{}. Table~\ref{tab:prior} summarizes these prior results. Recent direct-regression approaches trained on SPEED+ synthetic imagery, such as FastPose-ViT~\cite{fastposevit}, PAID-ViT~\cite{paidvit}, and Cov2Pose~\cite{cov2pose}, report mean rotation errors above 11\degree{} on the HIL domains and are not included in Table~\ref{tab:prior}. Recent keypoint-based approaches include mask-guided attention and adaptive keypoint suppression by Peng et al.~\cite{peng2026mgksnet} and a depth-aware keypoint loss by Yang et al.~\cite{yang2026a3dks}; neither study reports separate results on the full SPEED+ lightbox and sunlamp test sets.

\begin{table*}[t]
\caption{Selected prior published results on SPEED+ HIL. *~=~used real HIL imagery during training; \dag~=~3-seed ensemble. Values are reported under each source's own training and crop/detection protocol and are not a controlled comparison under one common pipeline. Bold denotes the best displayed value in each metric column, including rounded ties. Our rotation-error values in Table~\ref{tab:ours} are unthresholded means; the prior values in this table are reproduced as reported and may apply the HIL precision floor.}
\label{tab:prior}
\centering\small
\begin{tabular}{lcc ccc ccc}
\toprule
& & & \multicolumn{3}{c}{lightbox} & \multicolumn{3}{c}{sunlamp} \\
\cmidrule(lr){4-6}\cmidrule(lr){7-9}
Method & Year & Params & \et{} & \eq{}[$^\circ$] & \epose{} & \et{} & \eq{}[$^\circ$] & \epose{} \\
\midrule
TangoUnchained*~\cite{spec2021} & 2022 & --- & 0.018 & 3.19 & 0.073 & 0.015 & 4.30 & 0.090 \\
lava1302*~\cite{spec2021} & 2022 & --- & 0.048 & 6.66 & 0.165 & \textbf{0.011} & 2.73 & \textbf{0.059} \\
SPNv2 + ODR*~\cite{spnv2} & 2023 & 52.5M & 0.025 & 5.58 & 0.122 & 0.026 & 9.79 & 0.197 \\
Chen et al.*~\cite{chen2024} & 2024 & --- & 0.018 & 2.87 & 0.068 & 0.014 & 2.75 & 0.062 \\
EagerNet~\cite{eagernet} & 2023 & 88.6M & \textbf{0.009} & \textbf{1.75} & \textbf{0.039} & 0.013 & \textbf{2.66} & \textbf{0.059} \\
SPNv3-S~\cite{spnv3} & 2024 & 22.7M & 0.016 & 3.04 & 0.069 & 0.020 & 4.37 & 0.097 \\
SPNv3-S\dag~\cite{spnv3} & 2024 & 22.7M & 0.017 & 2.69 & 0.064 & 0.020 & 3.88 & 0.088 \\
SPNv3-B\dag~\cite{spnv3} & 2024 & 86.3M & 0.012 & 2.03 & 0.047 & 0.015 & 3.40 & 0.074 \\
Bergia et al.~\cite{bergia2026} (synthetic only, 40\% holdout) & 2026 & 303M & 0.012 & 3.27 & 0.069 & 0.016 & 4.06 & 0.087 \\
Bergia et al.*~\cite{bergia2026} (+ self-training, 40\% holdout) & 2026 & 303M & 0.012 & 2.57 & 0.057 & 0.015 & 3.25 & 0.072 \\
Bergia et al.*~\cite{bergia2026} (+ self-training, full set) & 2026 & 303M & 0.013 & 2.87 & 0.063 & 0.015 & 3.27 & 0.072 \\
\bottomrule
\end{tabular}
\end{table*}

Previous research has generally used small models compared to the large ViT foundation models increasingly used in other applications. Some models, such as SPNv3, were specifically designed around operating on limited compute. In the coming years, though, it is likely that onboard compute resources will become more common and less cost prohibitive~\cite{hpsc}. We set out to answer whether scaling improves pose estimation and to measure the computational cost of the resulting model on embedded hardware. Our research shows that there is indeed significant headroom and that previous architectures, when adapted to use cutting edge SSL vision foundation models, produce state of the art mean rotation errors on the SPEED+ lightbox and sunlamp test sets that improve on all previously published results we know of for those domains.

\section{Method}

We began by implementing the SPNv3 architecture described by Park and D'Amico in ``Bridging the Domain Gap for Flight-Ready Spaceborne Vision''~\cite{spnv3}. SPNv3 is a three stage architecture involving:
\begin{enumerate}
\item A bounding box detection algorithm that crops the high resolution input image to produce a smaller image of the target object suitable to be fed into the subsequent vision model.
\item Detection of $K$ predefined keypoints on the image of the target
\item A pose solution solver that takes the 2D keypoint locations and produces a 6D pose solution using the well-established Perspective-n-Point (PnP) family of solvers~\cite{epnp}.
\end{enumerate}

For our purposes, we can more or less treat steps 1 and 3 of this architecture as fixed. Following SPNv3, we use ground-truth-derived crops in place of a bounding box detection algorithm for stage 1 so that our comparison isolates the keypoint detection task central to stage 2.\footnote{Our SPEED+ results therefore evaluate the keypoint-and-PnP stage given a ground-truth-derived crop, the same protocol as SPNv3. Comparisons with methods that include their own detector, such as EagerNet and the SPEC2021 entries, are not under an identical protocol.} The pose solution solving math in step 3 is already well established.\footnote{During training, crops are randomly enlarged and shifted to build robustness against imperfect bounding boxes, as in SPNv3.} Remaining error in pose solutions for this architecture stem from inaccuracies in placing keypoints in stage 2.

In the case of SPNv3, stage 2 uses a neural net feeding heads that output $K$ heatmaps for the keypoints ($K=11$ for Tango). SPNv3 trained ViTPose, originally proposed by Xu et al.~\cite{vitpose}, for this purpose. For our implementation, we decided to tune DINOv3~\cite{dinov3}, a self-supervised vision foundation model by Meta AI, in place of ViTPose's ViT-S backbone. We trained the 300M variant using both LoRA~\cite{lora} and DoRA~\cite{dora}, and the 840M variant using LoRA, across only synthetic imagery. As with previous implementations, we found that random seeding multiple models and predicting via consensus between an ensemble of 3 produced superior results for the 300M model. For the rank-64 840M model, the three independently trained models with four-rotation test-time augmentation (the crop rotated by 0\degree{}, 90\degree{}, 180\degree{}, and 270\degree{}, with the back-rotated heatmaps averaged using each view's peak confidence as its weight) gave mean rotation errors of 1.19--1.20\degree{} on lightbox and 1.49--1.79\degree{} on sunlamp. Combining the three models with the same augmentation gave 1.17\degree{} and 1.56\degree{}, respectively. We report the single-model and ensemble configurations separately because they require four and twelve network passes per frame.

Our training, as mentioned, was purely on the official synthetic training split of SPEED+ (approximately 48{,}000 renders), with the 11{,}994-image synthetic validation split held out. Across all rank-16 training runs our configuration was held constant with the exception of the encoder itself and per-encoder input normalization matched to the encoder's specific pre-training statistics. The rank-64 runs raise the adapter rank from 16 to 64 (with LoRA scaling $\alpha/r$ of 2 instead of 1), and the rank-64 840M runs additionally lower the learning rate to $3\times10^{-4}$ for training stability. Network inputs were always the same $448\times448$ crops. The heatmap head was trained from scratch in every run (two transposed-convolution upsampling blocks and a $1\times1$ convolution producing $112\times112$ heatmaps, trained with a mean-squared-error loss against Gaussian targets). The original optimization schedule used AdamW with learning rate $10^{-3}$, weight decay $10^{-4}$, 30 epochs at batch size 32, and one warmup epoch followed by cosine decay; the rank-64 840M runs retain this schedule with the lower learning rate stated above. Thanks to modern fine-tuning methods such as LoRA and DoRA we were able to achieve our results while adding trainable parameters equivalent to roughly 0.5--3\% of total encoder weights in each tuning run, with the original encoder weights remaining frozen; adapters are placed on the query/key/value and output projections of every attention block. Since SPEED+ does not provide a real-image validation split, we follow the standards set in prior work and use the lightbox and sunlamp sets both to compare settings and to choose inference-time settings. No real images were used for training. As an additional check on our results, we used only the even-numbered frames of each real set to choose the inference-time settings and then scored the odd-numbered frames with the configuration selected that way, a single model with four-rotation test-time augmentation. This resulted in mean rotation errors of 1.22\degree{} on lightbox and 1.52\degree{} on sunlamp, consistent with the full-set results.

\section{Controls}

\begin{table*}[t]
\caption{Our results. Select State of the Art results are rendered in grey italic rows for comparison. Flip\% = fraction of frames with rotation error above $30^\circ$, i.e.\ a catastrophic orientation error; on this nearly symmetric target these are frequently the mirrored pose. Single-model rows are seed 2026, the first trained; the three-seed summaries average seeds 2026, 2027, and 2028 for the indicated one-pass or four-pass configuration, with sample standard deviation reported for rotation error. TTA = confidence-weighted average of 4 rotated passes. \ddag~=~lightbox translation means are dominated by a single physically invalid PnP solution with a translation error of 500\,m or more. \epose{} follows the revised SPEED+ definition used in SPNv3.}
\label{tab:ours}
\centering\small
\setlength{\tabcolsep}{4pt}
\begin{tabular}{lcc cccc cccc}
\toprule
& & & \multicolumn{4}{c}{lightbox} & \multicolumn{4}{c}{sunlamp} \\
\cmidrule(lr){4-7}\cmidrule(lr){8-11}
Model & Params & Passes & \et{} & \eq{}[$^\circ$] & flip\% & \epose{} & \et{} & \eq{}[$^\circ$] & flip\% & \epose{} \\
\midrule
\rowcolor{gray!12}\itshape EagerNet~\cite{eagernet} (reference) & \itshape 88.6M & \itshape 4 & \itshape 0.009 & \itshape 1.75 & --- & \itshape 0.039 & \itshape 0.013 & \itshape 2.66 & --- & \itshape 0.059 \\
\rowcolor{gray!12}\itshape SPNv3-B\dag~\cite{spnv3} (reference) & \itshape 86.3M & \itshape 3 & \itshape 0.012 & \itshape 2.03 & --- & \itshape 0.047 & \itshape 0.015 & \itshape 3.40 & --- & \itshape 0.074 \\
\midrule
DINOv3-L, LoRA r16 & 303M & 1 & 0.0099 & 1.90 & 0.43 & 0.0428 & 0.0117 & 2.26 & 0.47 & 0.0510 \\
\;\;L r16 + 3-seed ensemble & 303M & 3 & 0.0096 & 1.69 & 0.39 & 0.0387 & 0.0115 & 2.17 & 0.36 & 0.0492 \\
\;\;L r16 + 4-rot TTA & 303M & 4 & 0.0124 & 1.72 & 0.49 & 0.0421 & 0.0139 & 2.01 & 0.32 & 0.0488 \\
\;\;L r16 + ensemble + TTA & 303M & 12 & \ddag & 1.56 & 0.31 & \ddag & 0.0122 & 2.00 & 0.32 & 0.0470 \\
DINOv3-L, DoRA r16 & 303M & 1 & \ddag & 1.79 & 0.33 & \ddag & 0.0136 & 2.35 & 0.36 & 0.0544 \\
\;\;L DoRA r16 + 4-rot TTA & 303M & 4 & 0.0097 & 1.59 & 0.30 & 0.0371 & 0.0124 & 2.15 & 0.39 & 0.0497 \\
\midrule
DINOv3-H+, LoRA r16 & 840M & 1 & 0.0082 & 1.34 & 0.13 & 0.0312 & 0.0108 & 1.84 & 0.21 & 0.0428 \\
\;\;H+ r16, 3-seed mean $\pm$ SD & 840M & 1 & 0.0082 & 1.41$\pm$0.07 & 0.25 & 0.0326 & 0.0109 & 1.87$\pm$0.06 & 0.27 & 0.0434 \\
\;\;H+ r16 + 3-seed ensemble & 840M & 3 & 0.0085 & 1.31 & 0.19 & 0.0309 & 0.0102 & 1.72 & 0.21 & 0.0400 \\
\;\;H+ r16 + 4-rot TTA & 840M & 4 & 0.0076 & 1.20 & 0.12 & 0.0281 & 0.0112 & 1.67 & 0.18 & 0.0401 \\
\;\;H+ r16 + ensemble + TTA & 840M & 12 & 0.0075 & 1.21 & 0.13 & 0.0283 & 0.0096 & 1.68 & 0.21 & 0.0388 \\
\midrule
DINOv3-L, LoRA r64 & 303M & 1 & \ddag & 1.46 & 0.19 & \ddag & 0.0117 & 1.94 & 0.32 & 0.0454 \\
\;\;L r64 + 4-rot TTA & 303M & 4 & 0.0098 & 1.35 & 0.18 & 0.0330 & 0.0121 & 1.71 & 0.21 & 0.0418 \\
\midrule
DINOv3-H+, LoRA r64 & 840M & 1 & 0.0085 & 1.41 & 0.19 & 0.0327 & 0.0099 & 1.77 & 0.29 & 0.0407 \\
\;\;H+ r64, 3-seed mean $\pm$ SD & 840M & 1 & 0.0084 & 1.37$\pm$0.04 & 0.21 & 0.0319 & 0.0099 & 1.82$\pm$0.06 & 0.29 & 0.0414 \\
\;\;H+ r64 + 3-seed ensemble & 840M & 3 & 0.0076 & 1.22 & 0.10 & 0.0286 & \textbf{0.0095} & 1.74 & 0.29 & 0.0398 \\
\;\;H+ r64 + 4-rot TTA (seed 2026) & 840M & 4 & 0.0077 & 1.20 & \textbf{0.09} & 0.0283 & 0.0105 & \textbf{1.49} & \textbf{0.14} & \textbf{0.0363} \\
\;\;H+ r64 + TTA, 3-seed mean $\pm$ SD & 840M & 4 & 0.0075 & 1.197$\pm$0.004 & 0.11 & 0.0281 & 0.0117 & 1.67$\pm$0.16 & 0.24 & 0.0407 \\
\;\;H+ r64 + ensemble + TTA & 840M & 12 & \textbf{0.0073} & \textbf{1.17} & 0.10 & \textbf{0.0273} & 0.0113 & 1.56 & 0.21 & 0.0383 \\
\bottomrule
\end{tabular}
\end{table*}

We conducted a number of control runs in an effort to confirm the gains we achieved were due to tuning a larger encoder and not a difference in our architecture or improvements generally in base encoders (Table~\ref{tab:controls}):

\begin{enumerate}
\item Since we followed SPNv3 closely architecturally, we conducted our own training run using a similar baseline encoder (22M DeiT3~\cite{deit3} ViT-S) and obtained results in the same range.
\item To test whether modern ViT encoders can support pose estimation without fine tuning, we attempted to train only the head while using DINOv3 and SigLIP2-L~\cite{siglip2} as encoders in their own respective runs. Each failed to produce usable results when the encoder was frozen for training (DINOv3: 18\% / 29\% catastrophic flips on lightbox / sunlamp; SigLIP2-L: 54\% / 70\%).
\item EUPE-S~\cite{eupe}, a modern 22M parameter encoder, was trained with LoRA to test whether modern encoders still produced similar results to similarly sized encoders used in previous models like SPNv3. EUPE-S with LoRA actually ended up producing significantly worse results compared to our fully trained 22M DeiT3~\cite{deit3} ViT-S baseline, showing that using modern encoders and fine-tuning techniques alone does not improve on previous results.
\item We trained SigLIP2-L using LoRA to confirm that other modern large pretrained ViT models show similar improvements with tuning and that our results are not unique to DINOv3's architecture. DINOv3 300M does seem to be a better model for this application, as SigLIP2-L (316M) remained 18--23\% behind DINOv3-L (303M) with LoRA (1.90\degree{} / 2.26\degree{} vs 2.47\degree{} / 2.74\degree{}).
\end{enumerate}

\begin{table*}[t]
\caption{Controls results. mean$^\circ$ = mean rotation error over all frames; bulk$^\circ$ = mean over frames with error at or below $30^\circ$, i.e.\ the conditional mean on non-flipped frames; flip\% as in Table~\ref{tab:ours}. All control rows use seed 2026 and one network pass per crop; the LoRA controls use rank 16.}
\label{tab:controls}
\centering\small
\begin{tabular}{lcc ccc ccc}
\toprule
& & & \multicolumn{3}{c}{lightbox} & \multicolumn{3}{c}{sunlamp} \\
\cmidrule(lr){4-6}\cmidrule(lr){7-9}
Model & Params & Trained & mean$^\circ$ & bulk$^\circ$ & flip\% & mean$^\circ$ & bulk$^\circ$ & flip\% \\
\midrule
Baseline (SPNv3-style) & 22M & all & 3.60 & 2.20 & 1.5 & 5.19 & 3.21 & 2.1 \\
DINOv3-L frozen & 303M & head & 15.47 & 3.65 & 18.1 & 24.49 & 5.01 & 29.2 \\
SigLIP2-L frozen & 316M & head & 44.37 & 7.35 & 53.9 & 58.92 & 10.18 & 70.4 \\
EUPE-S + LoRA & 22M & head+LoRA & 7.99 & 3.32 & 5.9 & 8.93 & 3.83 & 6.4 \\
SigLIP2-L + LoRA & 316M & head+LoRA & 2.47 & 1.83 & 0.7 & 2.74 & 2.19 & 0.5 \\
\bottomrule
\end{tabular}
\end{table*}

\section{Hardware Testing}

While not a primary focus of this paper, we did want to benchmark the performance of our system on hardware that has already flown real missions as a preliminary test to see if computation of this size is viable. This is especially important since many past approaches, including the SPNv3 approach on which we based our model, cite computation constraints as a reason for opting for a smaller model.

Bergia et al.~\cite{bergia2026} also evaluated DINOv3-L on a Jetson Orin NX at $224\times224$ input using TensorRT FP16. Our measurements characterize the larger 840M model at $448\times448$ input, including measured board power.

We chose to test on an NVIDIA Jetson Orin NX 16GB, a $70\times45$\,mm system-on-module with a 1{,}792-core Ampere GPU, 16\,GB of shared memory, and a configurable 10--40\,W power envelope~\cite{orinnx} from a module family flown in 2026 on Aethero's Phobos~\cite{aethero2026} and EDGX's STERNA~\cite{nvidiaspace} payloads. Our preliminary findings are positive, showing that our 840M model is able to run at 133.8\,ms per single network pass at 32.0\,W measured board draw, and 419.8\,ms at 10.8\,W; four-view batches take 501.9\,ms and 1.63\,s at those respective operating points.

\begin{table}[t]
\caption{Embedded network benchmark on Jetson Orin NX 16GB, DINOv3-H+ with folded rank-64 LoRA, seed 2026, TensorRT BF16, $448\times448$ inputs. Latencies are median GPU compute times over 200 timed executions after warm-up. Batch 1 is one network pass; batch 4 measures a batch of four views, the network workload for single-model rotation TTA. Batch-4 latency is per batch, not per view. These timings exclude acquisition, crop preparation, transfers, TTA aggregation, and pose solving. Timing runs use random input tensors; agreement with the PyTorch evaluation was checked separately on real crops. Power is measured total board input draw, distinct from the named power mode; RAM is observed peak system memory, including runtime overhead.}
\label{tab:jetson}
\centering\small
\setlength{\tabcolsep}{3.5pt}
\begin{tabular}{l rr rr r}
\toprule
& \multicolumn{2}{c}{latency [ms]} & \multicolumn{2}{c}{board draw [W]} & peak \\
\cmidrule(lr){2-3}\cmidrule(lr){4-5}
Power mode & Batch 1 & Batch 4 & idle & load & RAM [GB] \\
\midrule
MAXN\_SUPER & 133.8 & 501.9 & 7.8 & 32.0 & 5.4 \\
40\,W & 133.1 & 502.3 & 6.7 & 32.4 & 6.2 \\
25\,W & 360.6 & 1343 & 6.8 & 13.4 & 6.6 \\
15\,W & 410.5 & 1597 & 6.7 & 12.7 & 5.3 \\
10\,W & 419.8 & 1630 & 4.9 & 10.8 & 6.0 \\
\bottomrule
\end{tabular}
\end{table}

It should be noted that the Orin, while small enough to be suitable for some missions, is a heavier module, in power and thermal terms, than is often needed or flown in many cases (its rated envelope is 10--40\,W, and we measured 10.8--32\,W of total board draw). Additionally, we have not tested the capabilities of the chip under radiation or in a flight-representative thermal environment. Given these limitations, these results are informative for assessing viability in cases where a mission is already being flown with sufficient compute on board. More testing is needed for true flight-ready assessment and for viability on smaller, more energy-constrained craft.

\section{Results}

\begin{figure}[t]
\centering
\includegraphics[width=0.9\columnwidth]{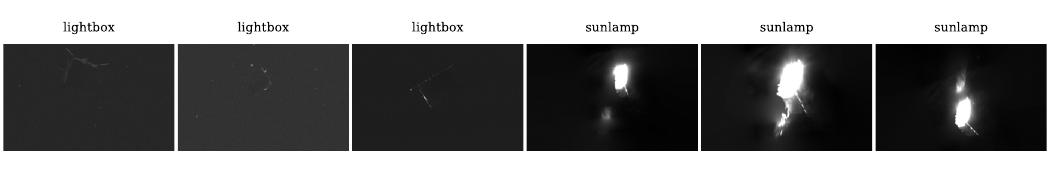}
\caption{Example frames which fail for two training seeds of the 300M DINOv3-L + LoRA model and for the 22M baseline: near-black lightbox frames (left) and glare-saturated sunlamp frames (right). Imagery from SPEED+~\cite{speedplus}, CC BY-NC-SA 4.0.}
\label{fig:failures}
\end{figure}

\begin{figure}[t]
\centering
\includegraphics[width=0.7\columnwidth]{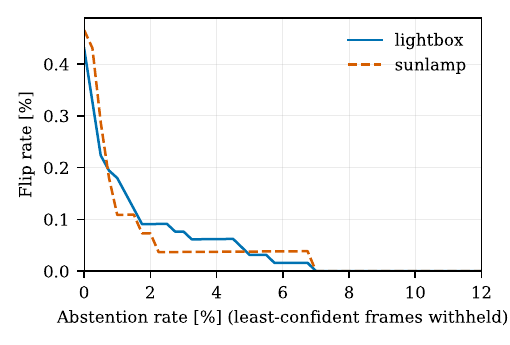}
\caption{Percent of the retained real frames the model incorrectly flipped, at various rates of least-confident frame abstention, for the 300M DINOv3-L + LoRA model.}
\label{fig:gate}
\end{figure}

Our rank-16 300M model in a single pass (1.90\degree{} lightbox / 2.26\degree{} sunlamp) achieves lower mean rotation errors than every previously published single-pass model result we know of on the full lightbox and sunlamp test sets (best prior single model, SPNv3-B: 2.18\degree{} / 3.61\degree{}), and with 4-rotation test-time augmentation (1.72\degree{} / 2.01\degree{}) it also exceeds the next best result, EagerNet, with four network passes, matching EagerNet's rotation count (1.75\degree{} / 2.66\degree{}). Our rank-16 840M model in a single pass (1.34\degree{} / 1.84\degree{}) exceeds every rank-16 multi-pass configuration of the 300M model and all previously published mean rotation-error results we know of on the full lightbox and sunlamp test sets, with or without ensemble or test-time augmentation (Table~\ref{tab:ours}).

With rank-64 adaptation, the 840M three-seed ensemble and four-rotation test-time augmentation gives 1.17\degree{} mean rotation error on lightbox and 1.56\degree{} on sunlamp, with catastrophic-flip rates of 0.10\% and 0.21\% and pose scores of 0.0273 and 0.0383, respectively. These results are computed over the full test sets without abstention. This configuration uses twelve network passes per frame; the Jetson measurements separately characterize one-pass and four-pass inference for a single model.

For the rank-64 840M ensemble with four-rotation test-time augmentation, median translation error is 2.5\,cm on lightbox and 3.4\,cm on sunlamp, at target distances of 2.5--9.6\,m. The ensemble fails on 13 of the 9{,}531 real frames, and one frame in each domain fails for every training seed. These are often near-black or glare-saturated (Fig.~\ref{fig:failures}). Many of these failures are flips that result in an extreme incorrect pose estimation in excess of 30 degrees. These flips often happen when the model mirrors the locations of the keypoints on the largely symmetric Tango satellite. A range check on the solved distance would remove the \ddag{} cases in Table~\ref{tab:ours}, but it catches only one flip per domain while the other flips solve to a plausible distance.

For the original rank-16 single-pass models, we observed that the model's confidence was low in most cases when catastrophic flips occurred. The 300M model showed a median per-frame confidence (the mean of its $K$ per-keypoint heatmap peak values) of 0.90 on correct frames vs 0.39 on flipped frames in the lightbox domain and 0.88 vs 0.41 in the sunlamp domain; the 840M model gives 0.92 vs 0.31 and 0.91 vs 0.39 on the respective domains. When broken down by degrees of error with and without flips included, our 300M model has a mean rotation error of 1.90\degree{} / 2.26\degree{} (lightbox / sunlamp) over all frames but 1.51\degree{} / 1.82\degree{} over the non-flipped frames alone; the 840M model 1.34\degree{} / 1.84\degree{} versus 1.23\degree{} / 1.52\degree{}. As can be seen in Fig.~\ref{fig:gate}, the model generally gives much lower confidence to flipped predictions. By retrospectively withholding the 2\% least-confident frames, the 300M model's flip rate falls from 0.43\% / 0.47\% to 0.09\% / 0.07\% and its mean rotation error to 1.46\degree{} / 1.81\degree{}. It is possible to remove every flip in both real domains by withholding 10\% of frames which would lower the mean error to 1.24\degree{} / 1.62\degree{}; the 840M model needs only 2\% withheld to reach zero flips in both domains (from 0.13\% / 0.21\%, against 1.5\% / 2.1\% for our fully trained 22M baseline), lowering its mean error from 1.34\degree{} / 1.84\degree{} to 1.16\degree{} / 1.47\degree{}, at the cost of introducing occasional blank responses between prediction frames. Our results tables do not employ this technique in order to remain comparable to previous methods, which report over the full test sets.

\section{Conclusion}

This paper shows that adapting large self-supervised ViT models such as DINOv3 improves on the results of previous methods that use relatively smaller encoders while holding other architecture elements constant. We find that these large encoders do not perform well when frozen but that modern fine tuning methods like LoRA and DoRA allow for adaptation of a small percent of encoder weights to achieve state of the art capabilities in this domain. This result suggests that future missions that require onboard pose estimation of a target craft, and that have available compute for a model of this size (on the Jetson Orin NX measured here, one network pass in 134--420\,ms at 11--32\,W of board draw), have a more accurate option than the previous state of the art. Additionally, our rank-16 results show that scaling from 300M to 840M cuts mean rotation error by roughly 15--30\% and roughly halves catastrophic flips, suggesting there may still be headroom to improve results further with larger foundation models. Our model was only trained and tested on a single target satellite with a well-defined 3D model and predefined keypoints, a stronger geometric prior than approximate-model methods such as EagerNet require, and limited to predictions from single frames.

\bibliographystyle{IEEEtran}
\bibliography{refs}

\ifanonymous\else
\par\medskip
{\footnotesize
\noindent\textbf{Author information}\par
\noindent\textbf{\authorA}\par
\noindent\expandafter\url\expandafter{\emailA}\par
\noindent\expandafter\url\expandafter{\orcidA}\par
\smallskip
\noindent\textbf{\authorB}\par
\noindent\expandafter\url\expandafter{\emailB}\par
\noindent\expandafter\url\expandafter{\orcidB}\par
}
\fi

\end{document}